\documentclass[conference]{IEEEtran}
\IEEEoverridecommandlockouts
\usepackage{cite}
\usepackage{amsmath,amssymb,amsfonts}
\usepackage{algorithmic}
\usepackage{graphicx}
\usepackage{textcomp}
\usepackage[linesnumbered,ruled,vlined]{algorithm2e}
\usepackage{graphicx}
\usepackage{subcaption}
\usepackage{graphicx}  
\usepackage{subcaption}  
\usepackage{comment}
\usepackage{algorithmic}
\usepackage{graphicx}
\usepackage{textcomp}
\usepackage{hyperref}  
\usepackage{amsmath}
\usepackage{tabularx}
\usepackage{booktabs}  

\usepackage{xcolor}
\usepackage{caption}
\usepackage{subcaption}
\usepackage{tabularx} \usepackage{xcolor}
\def\BibTeX{{\rm B\kern-.05em{\sc i\kern-.025em b}\kern-.08em
    T\kern-.1667em\lower.7ex\hbox{E}\kern-.125emX}}
\begin{document}

\title{
Prototype-Guided Diffusion: Visual Conditioning without External Memory
}

\author{\IEEEauthorblockN{1\textsuperscript{st} Bilal Faye}
faye@lipn.univ-paris13.fr
\and
\IEEEauthorblockN{2\textsuperscript{nd} Hanane Azzag}
azzag@univ-paris13.fr
\and
\IEEEauthorblockN{3\textsuperscript{rd} Mustapha Lebbah}
mustapha.lebbah@uvsq.fr}
\maketitle

\begin{abstract}
Diffusion models achieve state-of-the-art image generation but remain computationally costly due to iterative denoising. Latent-space models like Stable Diffusion reduce overhead yet lose fine detail, while retrieval-augmented methods improve efficiency but rely on large memory banks, static similarity models, and rigid infrastructures.\newline
We introduce the \textit{Prototype Diffusion Model (PDM)}, which embeds prototype learning into the diffusion process to provide adaptive, memory-free conditioning. Instead of retrieving references, PDM learns compact visual prototypes from clean features via contrastive learning, then aligns noisy representations with semantically relevant patterns during denoising. Experiments demonstrate that PDM sustains high generation quality while lowering computational and storage costs, offering a scalable alternative to retrieval-based conditioning.
\end{abstract}


\section{Introduction}
Diffusion models have rapidly become a dominant generative framework, known for high sample quality, training stability, and flexibility across tasks. Since the introduction of Denoising Diffusion Probabilistic Models (DDPMs)~\cite{ho2020denoising}, they have surpassed GAN-based approaches~\cite{maze2023diffusion,dhariwal2021diffusion} and been extended to unconditional generation~\cite{song2020denoising}, super-resolution~\cite{saharia2022photorealistic}, inpainting~\cite{lugmayr2022repaint}, and text-to-image generation~\cite{zhou2023shifted,rombach2022high}.\newline
Despite these advances, diffusion models remain slow due to iterative denoising. Latent-space methods such as Stable Diffusion~\cite{rombach2022high} mitigate this cost but lose spatial detail and require powerful encoders/decoders. Retrieval-augmented approaches (RDM)~\cite{blattmann2022retrieval} improve efficiency by conditioning on examples retrieved via pretrained vision-language models (e.g., CLIP~\cite{radford2021learning}), but incur large memory/storage costs and rely on fixed features, limiting adaptability. An orthogonal direction, ProtoDiffusion~\cite{baykal2024protodiffusion}, introduces prototype-based conditioning but in a two-stage pipeline: class-specific prototypes are precomputed and remain static, hindering adaptability and scalability to fine-grained or multimodal settings.\newline
We address these limitations with the \textit{Prototype Diffusion Model (PDM)}, which integrates online prototype learning directly into the diffusion process. Unlike ProtoDiffusion, PDM requires no labels and is fully unsupervised: prototypes are dynamically constructed from clean intermediate features using contrastive objectives, then reused as semantic anchors during noisy denoising steps. This joint, continuous learning allows prototypes to adapt alongside evolving representations, improving semantic grounding and efficiency without external memory. Furthermore, when labels are available, we extend PDM to a supervised variant, \textit{s-PDM}, where class-specific prototypes are fixed by labels, eliminating prototype selection and compactness loss. Our contributions are:
\begin{itemize}
    \item We propose \textit{PDM}, the first diffusion model that integrates online, unsupervised prototype learning into the denoising process, enabling efficient and adaptive conditioning without external memory or annotations.
    \item We extend PDM to a supervised setting with \textit{s-PDM}, where label-driven prototypes further simplify training and remove the need for compactness regularization.
    \item Through extensive experiments, we show that both PDM and s-PDM outperform standard DDPMs (no prototypes) and ProtoDiffusion (two-stage static prototypes), validating their scalability and effectiveness.
\end{itemize}

\section{Related Work}

\subsection{Diffusion Models}
Diffusion models have become a dominant generative framework, offering strong performance in image synthesis, inpainting, and super-resolution since DDPMs~\cite{ho2020denoising}. Latent diffusion~\cite{rombach2022high} improves efficiency by operating in compressed spaces, while recent advances such as consistency models~\cite{song2023} and SDXL~\cite{podell2024} demonstrate scalability. However, the denoising process remains slow due to many sequential steps, and latent-space acceleration often sacrifices spatial fidelity. These limitations motivate methods that incorporate additional context to guide denoising.

\subsection{Retrieval-Augmented Diffusion Models}
Retrieval-based approaches improve quality and controllability by conditioning diffusion on externally retrieved examples. RDM~\cite{blattmann2022retrieval} aligns noisy features with CLIP-based retrieved patches, and IRDiff~\cite{huang2024interaction} applies similar ideas to molecular generation. While effective, these methods rely on large static memory banks, frozen encoders~\cite{radford2021learning}, and incur retrieval latency, reducing adaptability to evolving representations. Such constraints motivate lighter, internal conditioning mechanisms.

\subsection{Prototype Learning in Diffusion Models}
Prototype-based conditioning offers compact alternatives to external memory. ProtoDiffusion~\cite{baykal2024protodiffusion} employs class-specific prototypes but learns them in a separate, static stage, limiting adaptability. Other works use prototypes for few-shot or continual learning~\cite{du2024protodiff,doan2023class}, yet typically rely on coarse class-level structures or focus on discriminative tasks. Existing methods thus lack fully joint, fine-grained prototype learning integrated into diffusion.

\smallskip
Our Prototype Diffusion Model (PDM) addresses this gap by learning patch-level prototypes online from clean features and aligning them with noisy representations during denoising, enabling adaptive conditioning without external memory.

\section{Background}

\subsection{Diffusion Models}

Diffusion models are a class of generative models that learn to synthesize data by inverting a stochastic noising process. A prominent and widely used variant is the \textit{Denoising Diffusion Probabilistic Model} (DDPM)~\cite{sohl2015deep, ho2020denoising}, which formulates generation as the reversal of a predefined forward diffusion process.\newline
The forward process gradually corrupts a clean data sample $x_0 \in \mathbb{R}^d$, drawn from an unknown data distribution $q(x_0)$, by adding Gaussian noise over $T$ discrete timesteps. This results in a sequence of latent variables $x_1, \dots, x_T$, defined recursively by:
\begin{equation}
    q(x_t \mid x_{t-1}) = \mathcal{N}(x_t; \sqrt{1 - \beta_t} \, x_{t-1}, \beta_t \, \mathbf{I}),
\end{equation}
where $\beta_t \in [0, 1]$ is a variance schedule controlling the amount of noise injected at each timestep, and $\mathbf{I}$ is the identity matrix.\newline
By defining $\alpha_t = 1 - \beta_t$ and the cumulative product $\bar{\alpha}_t = \prod_{s=1}^{t} \alpha_s$, the marginal distribution of $x_t$ given $x_0$ admits a closed-form expression:
\begin{equation}
    x_t = \sqrt{\bar{\alpha}_t} \, x_0 + \sqrt{1 - \bar{\alpha}_t} \, \epsilon, \quad \epsilon \sim \mathcal{N}(0, \mathbf{I}).
\end{equation}
The goal of the model is to learn the reverse process $p_\theta(x_{t-1} \mid x_t)$, which is intractable in general. DDPM approximates this reverse process by a parameterized Gaussian distribution:
\begin{equation}
    p_\theta(x_{t-1} \mid x_t) = \mathcal{N}(x_{t-1}; \mu_\theta(x_t, t), \Sigma_t),
\end{equation}
where the mean $\mu_\theta(x_t, t)$ is predicted by a neural network and the variance $\Sigma_t$ is typically fixed or learned.\newline
A common parameterization predicts the noise $\epsilon$ added during the forward process. The mean of the reverse distribution is then recovered as:
\begin{equation}
    \mu_\theta(x_t, t) = \frac{1}{\sqrt{\alpha_t}} \left( x_t - \frac{1 - \alpha_t}{\sqrt{1 - \bar{\alpha}_t}} \, \epsilon_\theta(x_t, t) \right),
\end{equation}
where $\epsilon_\theta(x_t, t)$ is the model's estimate of the noise component $\epsilon$ at timestep $t$.\newline
The model is trained to minimize the expected mean squared error between the true noise $\epsilon$ and the predicted noise $\epsilon_\theta$, resulting in the simplified training objective:
\begin{equation}
    \mathcal{L}_\text{simple} = \mathbb{E}_{t, x_0, \epsilon} \left[ \left\| \epsilon - \epsilon_\theta\left( \sqrt{\bar{\alpha}_t} \, x_0 + \sqrt{1 - \bar{\alpha}_t} \, \epsilon, t \right) \right\|^2 \right].
\end{equation}
This objective encourages the model to accurately denoise corrupted inputs at arbitrary timesteps, ultimately enabling sampling by iteratively denoising from pure Gaussian noise $x_T \sim \mathcal{N}(0, \mathbf{I})$ back to a coherent data sample $x_0$.

\subsection{Retrieval-Augmented Diffusion Models}
Retrieval-Augmented Diffusion Models (RDMs) enhance classical diffusion models by incorporating a retrieval mechanism that conditions generation on external, real examples. This semi-parametric approach supplements the learned generative model with non-parametric memory, enabling improved sample diversity, controllability, and generalization~\cite{rombach2022high,blattmann2022retrieval,hu2023reveal}.\newline
Given an input image $x$, its latent representation $z = E(x)$ is computed using a pretrained encoder $E$, typically from an autoencoder such as VQ-GAN~\cite{esser2021taming}. A forward diffusion process is then applied in the latent space to produce noisy samples $z_1, \dots, z_T$, analogous to the DDPM framework.\newline
To enrich the generative process with semantic context, a retrieval function $\xi_k(x, D)$ selects the $k$ nearest neighbors of $x$ from a dataset $D$, based on similarity in a learned embedding space. This embedding space is induced by a pre-trained model $\phi$, such as CLIP~\cite{radford2021learning}, and the retrieval is defined as:
\begin{equation}
    \xi_k(x, D) = \text{Top-}k \left( \text{sim}(\phi(x), \phi(y)) \mid y \in D \right),
\end{equation}
where $\text{sim}(\cdot, \cdot)$ denotes a similarity metric, such as cosine similarity, in the embedding space $\phi(\cdot)$.\newline
The denoising model is then conditioned not only on the noisy latent $z_t$ and timestep $t$, but also on the retrieved support set $\{ \phi(y) \mid y \in \xi_k(x, D) \}$. The denoising function becomes:
\begin{equation}
    \epsilon_\theta(z_t, t, \{ \phi(y) \}),
\end{equation}
where conditioning is typically implemented using cross-attention layers that attend to the retrieved features during noise prediction~\cite{rombach2022high}.\newline
The training objective follows the denoising loss paradigm, extended to include retrieval conditioning:
\begin{equation}
    \mathcal{L}_\text{RDM} = \mathbb{E}_{x, \epsilon, t} \left[ \left\| \epsilon - \epsilon_\theta\left(z_t, t, \{ \phi(y) \mid y \in \xi_k(x, D) \} \right) \right\|^2 \right].
\end{equation}
This formulation enables the model to ground its generation in real, semantically similar samples, which is particularly beneficial in data-scarce regimes or when fine control over the output is desired. Unlike purely parametric approaches, RDMs flexibly adapt to new domains by querying from an updated external memory without retraining the entire model.

\subsection{Prototype Learning in Diffusion Models}
Prototype-based learning has become a cornerstone of few-shot learning, where the objective is to generalize to novel classes from only a few labeled examples. One of the most influential approaches in this domain is the Prototypical Network (ProtoNet)~\cite{snell2017prototypical}, which embeds both support and query samples into a common metric space and performs classification based on distances to class prototypes.\newline
Given a set of $K$ support examples $\{x_{c,k}\}_{k=1}^K$ for a class $c$, a class prototype $z_c \in \mathbb{R}^d$ is computed as the mean of their embeddings under a shared encoder $f_\phi$:
\begin{equation}
    z_c = \frac{1}{K} \sum_{k=1}^{K} f_\phi(x_{c,k}),
\end{equation}
where $f_\phi$ is typically a neural network trained to produce discriminative embeddings. A query sample $x_q$ is classified by computing the softmax over negative distances to each prototype:
\begin{equation}
    p(y = c \mid x_q) = \frac{\exp(-d(f_\phi(x_q), z_c))}{\sum_{c'} \exp(-d(f_\phi(x_q), z_{c'}))},
\end{equation}
where $d(\cdot, \cdot)$ is usually the squared Euclidean distance in the embedding space.
In the context of diffusion models, class prototypes can be leveraged to guide the generative process, especially in scenarios where class-conditioning is required with limited labeled data. Instead of conditioning the diffusion process on textual prompts or class labels alone, prototype-guided diffusion injects semantic structure directly from few-shot support examples.\newline
Let $z_c$ denote the prototype for the target class $c$, and let $x_t$ be a noisy sample at timestep $t$. The denoising model can be conditioned on $z_c$ to produce a prototype-guided noise estimate $\epsilon_\theta(x_t, t, z_c)$. Following the classifier-free guidance strategy~\cite{ho2022classifier}, the final guided noise estimate is computed as:
\begin{equation}
    \epsilon_{\text{guided}}(x_t, t) = (1 + s) \cdot \epsilon_\theta(x_t, t, z_c) - s \cdot \epsilon_\theta(x_t, t),
\end{equation}
where $\epsilon_\theta(x_t, t)$ is the unconditional noise prediction, and $s \geq 0$ is a guidance scale parameter controlling the strength of conditioning. When $s = 0$, the model generates unconditionally; larger values of $s$ encourage outputs more aligned with the class prototype.\newline
This formulation enables the diffusion model to generate samples that are consistent with the semantic identity of a class, even in few-shot or low-resource regimes. Moreover, prototype guidance introduces minimal overhead while providing strong inductive bias via non-parametric, data-driven class priors.

\section{Method}
The Prototype Diffusion Model (PDM) is a generative model that leverages prototype learning to condition the denoising process in diffusion models. Unlike previous methods such as ProtoDiffusion, which rely on labeled datasets for supervised prototype assignment, PDM is trained in a fully unsupervised manner. This allows the model to discover and utilize semantic prototypes without the need for annotations, making it particularly well-suited for unlabeled data.\newline
The model is composed of two jointly trained neural networks. The first, denoted $f_{\phi}$, is a convolutional neural network responsible for feature extraction. Given an input image $x \in \mathbb{R}^{H \times W \times C}$, it produces a latent representation:
\begin{equation}
\hat{x} = f_{\phi}(x), \quad \hat{x} \in \mathbb{R}^{D}
\end{equation}
A set of $K$ prototypes $\{ e_1, \dots, e_K \} \subset \mathbb{R}^{D}$ is maintained as learnable parameters, randomly initialized at the start of training. For each image $x$, the closest prototype is selected based on Euclidean distance in the latent space:
\begin{equation}
e_x = \arg\min_{e_i} \left\| f_{\phi}(x) - e_i \right\|_2
\end{equation}
The second component of PDM is the denoising network $f_{\theta}$, implemented as a U-Net~\cite{ronneberger2015u}, which takes as input a noisy image and is conditioned on the selected prototype. \textit{To incorporate the prototype into the U-Net, cross-attention layers are added such that the prototype $e_x$ acts as both key and value, while the query originates from the output of the previous block processing the noisy image.} This enables the network to guide the denoising process using the most relevant semantic concept.\newline
Following the standard diffusion framework, Gaussian noise is added to the input image over $T$ steps using the forward process:
\begin{equation}
\tilde{x}_t = \sqrt{\bar{\alpha}_t} x + \sqrt{1 - \bar{\alpha}_t} \epsilon, \quad \epsilon \sim \mathcal{N}(0, I)
\end{equation}
At each time step $t$, the denoising network receives $\tilde{x}_t$, the selected prototype \( e_x \), and a time embedding $\gamma(t)$. The prototype is enriched with this temporal information before being passed to the attention mechanism:
\begin{equation}
\hat{\epsilon} = f_{\theta}(\tilde{x}_t, e_x + \gamma(t))
\end{equation}
The training objective of PDM includes several loss components. First, a contrastive loss encourages the feature extractor $f_{\phi}$ to bring the latent representation $f_{\phi}(x)$ close to its associated prototype while pushing it away from others:
\begin{equation}
\mathcal{L}_{\text{contrastive}} = -\log \left( \frac{\exp(-\tau \| f_{\phi}(x) - e_x \|_2^2)}{\sum_{k=1}^{K} \exp(-\tau\| f_{\phi}(x) - e_k \|_2^2)} \right)
\end{equation}
In addition, an alignment loss directly penalizes the squared Euclidean distance between the image representation and the selected prototype:
\begin{equation}
\mathcal{L}_{\text{align}} = \left\| f_{\phi}(x) - e_x \right\|_2^2
\end{equation}
To ensure that the learned prototypes are diverse and semantically meaningful, a compactness regularization term is introduced. This term penalizes high similarity between different prototypes using cosine similarity:
\begin{equation}
\mathcal{L}_{\text{compact}} = \sum_{k \neq k'} \text{sim}(e_k, e_{k'})
\end{equation}
where $\text{sim}(e_k, e_{k'})$ denotes the cosine similarity between prototypes $e_k$ and $e_{k'}$. This encourages the prototypes to capture distinct concepts rather than collapsing into similar representations.\newline
The diffusion objective follows the standard DDPM loss, measuring the mean squared error between the predicted noise and the true noise added at each time step:
\begin{equation}
\mathcal{L}_{\text{diff}} = \mathbb{E}_{x, t, \epsilon} \left[ \left\| \epsilon - f_{\theta}(\tilde{x}_t, e_x + \gamma(t)) \right\|_2^2 \right]
\end{equation}
The overall training loss for PDM combines all components:

\begin{equation}
\mathcal{L}_{\text{PDM}} = \mathcal{L}_{\text{diff}} + \mathcal{L}_{\text{contrastive}} + \mathcal{L}_{\text{align}} +  \mathcal{L}_{\text{compact}}
\end{equation}
The training procedure for the Prototype Diffusion Model is detailed in Algorithm~\ref{alg:pdm_training}.
\begin{algorithm}[!h]
\caption{Training Prototype Diffusion Model (PDM)}
\label{alg:pdm_training}
\KwIn{Dataset $\mathcal{D} = \{x_i\}_{i=1}^N$, number of prototypes $K$, noise schedule $\{\alpha_t\}_{t=1}^T$, weight $\tau$}
\KwOut{Trained networks $f_{\phi}, f_{\theta}$, and prototypes $\{e_1, \ldots, e_K\}$}

Initialize feature extractor $f_{\phi}$, denoiser $f_{\theta}$, and prototypes $\{e_1, \ldots, e_K\} \subset \mathbb{R}^D$\;

\ForEach{minibatch $\{x_1, \ldots, x_B\} \subset \mathcal{D}$}{

    \ForEach(\tcp*[f]{Feature encoding and prototype assignment}){$x_i \in \{x_1, \ldots, x_B\}$}{
        Compute $\hat{x}_i = f_{\phi}(x_i)$\;
        Select prototype: $e_{x_i} = \arg\min_{e_k} \| \hat{x}_i - e_k \|_2$\;
        Sample $t \sim \mathcal{U}\{1, T\},\ \epsilon \sim \mathcal{N}(0, I)$\;
        Compute $\bar{\alpha}_t = \prod_{s=1}^t \alpha_s$\;
        Compute $\tilde{x}_i = \sqrt{\bar{\alpha}_t} x_i + \sqrt{1 - \bar{\alpha}_t} \epsilon$\;
        Predict $\hat{\epsilon}_i = f_{\theta}(\tilde{x}_i, e_{x_i} + \gamma(t))$\;
    }

    Compute loss terms:

    $\mathcal{L}_{\text{contrastive}} = -\frac{1}{B} \sum_{i=1}^B \log \left( \frac{\exp(-\tau\| \hat{x}_i - e_{x_i} \|_2^2)}{\sum_{k=1}^K \exp(-\tau\| \hat{x}_i - e_k \|_2^2)} \right)$\;

    $\mathcal{L}_{\text{align}} = \frac{1}{B} \sum_{i=1}^B \| \hat{x}_i - e_{x_i} \|_2^2$\;

    $\mathcal{L}_{\text{compact}} = \beta \sum_{k \neq k'} \text{sim}(e_k, e_{k'})$\;

    $\mathcal{L}_{\text{diff}} = \frac{1}{B} \sum_{i=1}^B \| \epsilon_i - \hat{\epsilon}_i \|_2^2$\;

    Combine losses:

    $\mathcal{L}_{\text{PDM}} = \mathcal{L}_{\text{diff}} + \mathcal{L}_{\text{contrastive}} + \mathcal{L}_{\text{align}} +  \mathcal{L}_{\text{compact}}$\;

    Update $f_{\phi}, f_{\theta}, \{e_k\}$ via gradient descent on $\mathcal{L}_{\text{PDM}}$\;
}

\Return $f_{\phi}, f_{\theta}, \{e_k\}_{k=1}^K$\;
\end{algorithm}

During inference, PDM supports both conditional and unconditional generation. If an input image is provided, it is passed through $f_{\phi}$ to select the most appropriate prototype $e_x$, which is then used to condition the denoising process. If no image is provided, a prototype is selected at random. The complete inference process is summarized in Algorithm~\ref{alg:pdm_inference}.

\begin{algorithm}[!h]
\caption{Inference in Prototype Diffusion Model (PDM)}
\label{alg:pdm_inference}
\KwIn{Prototype set \( \{e_1, \ldots, e_K\} \), trained feature extractor \( f_{\phi} \), denoiser \( f_{\theta} \), optional image \( x \), noise schedule \( \{\alpha_t\}_{t=1}^T \)}
\KwOut{Generated image \( \hat{x}_0 \)}

\If{image \( x \) is provided}{
    Compute \( \hat{x} = f_{\phi}(x) \)\;
    Select prototype: \( e_x = \arg\min_{e_i} \| \hat{x} - e_i \|_2 \)\;
}
\Else{
    Sample prototype \( e_x \sim \{ e_1, \ldots, e_K \} \)\;
}

Initialize \( \tilde{x}_T \sim \mathcal{N}(0, I) \)\;

\For(\tcp*[f]{Iterative denoising}){$t \leftarrow T$ \KwTo $1$}{
    Compute \( \bar{\alpha}_t = \prod_{s=1}^t \alpha_s \)\;
    Compute \( \hat{\epsilon}_t = f_{\theta}(\tilde{x}_t, e_x + \gamma(t)) \)\;
    Compute \( \sigma_t = \sqrt{1 - \alpha_t} \), sample \( z \sim \mathcal{N}(0, I) \)\;
    
    Update:
    \[
    \tilde{x}_{t-1} = \frac{1}{\sqrt{\alpha_t}} \left( \tilde{x}_t - \frac{1 - \alpha_t}{\sqrt{1 - \bar{\alpha}_t}} \hat{\epsilon}_t \right) + \sigma_t z
    \]
}

\Return \( \hat{x}_0 = \tilde{x}_0 \)\;
\end{algorithm}

PDM can be readily adapted to a supervised context in which labeled data is available. We create and refer to this variant as \textit{supervised Diffusion Prototype Model (s-PDM)}. In this case, each class is associated with a fixed prototype. Therefore, during training, prototype selection is unnecessary since the correct prototype is known. Moreover, the compactness loss $\mathcal{L}_{\text{compact}}$ is no longer needed, as inter-class semantic separation is enforced by the labels themselves. The resulting supervised loss simplifies to:
\begin{equation}
\mathcal{L}_{\text{PDM-supervised}} = \mathcal{L}_{\text{diff}} + \mathcal{L}_{\text{contrastive}} + \mathcal{L}_{\text{align}}
\end{equation}
The inference procedure remains unchanged, conditioned either on a labeled image or a sampled prototype for unconditional generation. This flexibility allows PDM to operate seamlessly in both supervised and unsupervised settings.

\section{Experiments}

We evaluate PDM and its supervised variant s-PDM on CIFAR-10~\cite{krizhevsky2009learning}, STL-10~\cite{coates2011analysis}, EuroSAT~\cite{helber2019eurosat}, and Tiny ImageNet~\cite{le2015tiny} using two NVIDIA A100 GPUs. 
\subsection{Experimental Setup}
All experiments use a compact U-Net with channel configuration \texttt{[128, 256, 256, 256]}. Prototypes are injected at the bottleneck through a single cross-attention block, enabling global semantic conditioning with minimal computational cost. The feature extractor $f_\phi$ is a lightweight 4-layer CNN with adaptive average pooling. 

We compare our models to two baselines: (i) a standard DDPM (no prototypes) and (ii) ProtoDiffusion, where prototypes are learned in a separate stage and kept frozen. These comparisons allow us to isolate the effect of \emph{joint prototype learning during diffusion}, which is the primary novelty of PDM. Importantly, our goal is not to surpass state-of-the-art diffusion performance, but to demonstrate that \textbf{learning prototypes jointly with the denoiser leads to systematically improved generative quality}.

Performance is reported using Inception Score (IS)~\cite{salimans2016improved}, Fréchet Inception Distance (FID), and Kernel Inception Distance (KID)~\cite{binkowski2018demystifying}.

\subsection{Quantitative Evaluation}
\begin{table*}[!h]
\centering
\resizebox{\textwidth}{!}{
\begin{tabular}{l|ccc|ccc|ccc|ccc}
\toprule
\textbf{Method} & \multicolumn{3}{c|}{CIFAR-10} & \multicolumn{3}{c|}{STL-10} & 
\multicolumn{3}{c|}{EuroSAT} & \multicolumn{3}{c}{Tiny ImageNet} \\
& IS $\uparrow$ & FID $\downarrow$ & KID $\downarrow$
& IS $\uparrow$ & FID $\downarrow$ & KID $\downarrow$
& IS $\uparrow$ & FID $\downarrow$ & KID $\downarrow$
& IS $\uparrow$ & FID $\downarrow$ & KID $\downarrow$ \\
\midrule
DDPM           & 7.12 & 18.45 & 0.021 & 7.85 & 34.20 & 0.038 & 3.75 & 28.47 & 0.018 & 9.05 & 48.22 & 0.08 \\
ProtoDiffusion & \textbf{8.50} & 11.70 & 0.009 & 9.40 & 22.70 & 0.015 & 4.45 & 15.20 & 0.010 & 10.55 & 31.00 & 0.04 \\
PDM (unsup.)   & 8.35 & 8.10  & 0.007 & 10.28 & 21.30 & 0.011 & 4.40 & 13.50 & 0.008 & 11.20 & 25.40 & 0.03 \\
s-PDM (sup.)   & 8.20 & \textbf{6.58} & \textbf{0.004} & \textbf{11.77} & \textbf{20.32} & \textbf{0.007} & \textbf{4.52} & \textbf{11.29} & \textbf{0.005} & \textbf{11.65} & \textbf{23.00} & \textbf{0.02} \\
\bottomrule
\end{tabular}
}
\caption{Comparison of DDPM, ProtoDiffusion, PDM, and s-PDM across multiple datasets. Best values are in \textbf{bold}.}
\label{tab:results}
\end{table*}

Table~\ref{tab:results} shows that both PDM and s-PDM outperform the baselines across all datasets and metrics. The gains are especially clear for FID and KID, which require matching the real data distribution. 

\textbf{Key finding:} jointly learning prototypes within the diffusion process (PDM) consistently improves generative fidelity over DDPM and ProtoDiffusion. This confirms that prototypes must remain trainable and aligned with the evolving denoiser features; otherwise, as in ProtoDiffusion, they gradually lose relevance.

s-PDM obtains the strongest results overall due to label-informed prototype assignment. Its advantage is most visible on FID/KID, where class-consistent conditioning directly improves distributional accuracy. In contrast, IS does not rely on real samples and favors diversity, which slightly benefits the unsupervised PDM.

\begin{figure}[!h]
\centering
\includegraphics[width=0.45\textwidth]{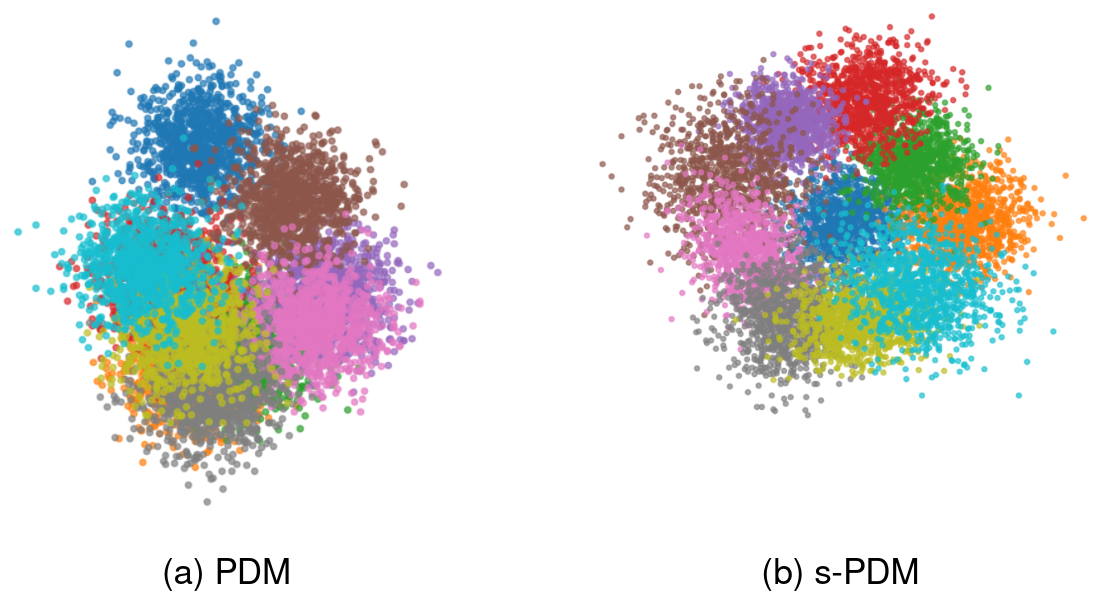}
\caption{PCA of $f_\phi$ features on CIFAR-10. s-PDM shows clearer class separation, while PDM forms semantic clusters without labels.}
\label{fig:clusters}
\end{figure}

Figure~\ref{fig:clusters} illustrates this difference: PCA of $f_\phi$ features reveals sharper class separation in s-PDM, while PDM forms compact but unlabeled semantic groups. This confirms that supervised prototypes enforce stronger structure in latent space.

\subsection{Effect of the Number of Prototypes}
To assess the importance of prototype granularity, we vary the number of prototypes in PDM while keeping all other settings fixed.

\begin{table}[!h]
\centering
\begin{tabular}{c|ccc}
\toprule
\textbf{Prototypes} & IS $\uparrow$ & FID $\downarrow$ & KID $\downarrow$ \\
\midrule
3  & 7.50 & 11.90 & 0.010 \\
5  & 7.85 & 10.20 & 0.009 \\
7  & 8.10 & 9.10  & 0.008 \\
10 & 8.35 & 8.10  & 0.007 \\
13 & 8.42 & 7.95  & 0.006 \\
15 & 8.48 & 7.85  & 0.006 \\
20 & 8.20 & 7.70  & 0.006 \\
\bottomrule
\end{tabular}
\caption{Ablation on the number of prototypes (CIFAR-10).}
\label{tab:prototype_variation}
\end{table}

Results (Table~\ref{tab:prototype_variation}) show that using fewer prototypes than the dataset's intrinsic semantic modes reduces performance, as multiple classes collapse into a single centroid. Increasing prototypes beyond the number of classes yields marginal gains but eventually causes over-fragmentation. These findings confirm that prototypes should match the dataset's semantic granularity—a desirable property for adaptive or continual generative modeling.

\subsection{Visualization of Generated Samples}

\begin{figure*}[!t]
    \centering
    \begin{subfigure}[t]{0.23\textwidth}
        \centering
        \includegraphics[width=\linewidth]{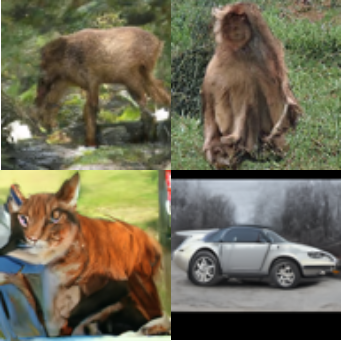}
        \caption{DDPM}
    \end{subfigure}
    \hspace{0.01\textwidth}
    \begin{subfigure}[t]{0.23\textwidth}
        \centering
        \includegraphics[width=\linewidth]{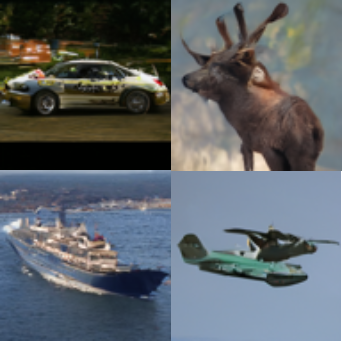}
        \caption{ProtoDiffusion}
    \end{subfigure}
    \hspace{0.01\textwidth}
    \begin{subfigure}[t]{0.23\textwidth}
        \centering
        \includegraphics[width=\linewidth]{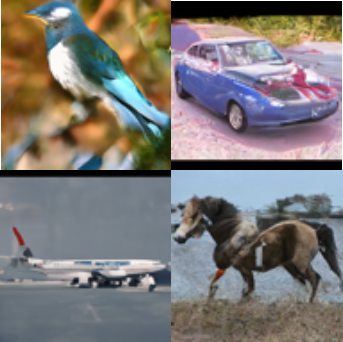}
        \caption{PDM (unsupervised)}
    \end{subfigure}
    \hspace{0.01\textwidth}
    \begin{subfigure}[t]{0.23\textwidth}
        \centering
        \includegraphics[width=\linewidth]{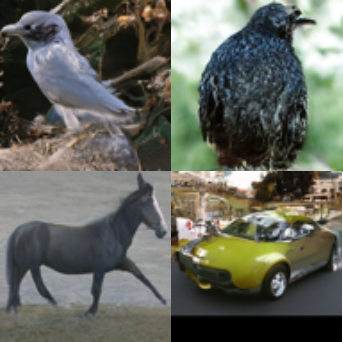}
        \caption{s-PDM (supervised)}
    \end{subfigure}

    \caption{Random $2\times2$ sample generations from models trained on STL-10 after 500 training epochs. Baseline models (DDPM and ProtoDiffusion) produce plausible but less semantically coherent images. Prototype-guided methods (PDM and s-PDM) demonstrate stronger semantic structure, with s-PDM further improving class consistency and visual fidelity. This highlights the impact of jointly learning prototypes on guiding the diffusion process.}
    \label{fig:generated_samples}
\end{figure*}

Figure~\ref{fig:generated_samples} presents randomly generated $2\times2$ image grids from four models trained on STL-10: DDPM, ProtoDiffusion, PDM, and s-PDM. All four models are shown side by side to emphasize differences in semantic coherence and prototype-guided generation.

These visualizations confirm that jointly learned prototypes substantially improve generation quality. While DDPM and ProtoDiffusion produce plausible images, they lack semantic consistency and class-specific alignment. PDM, even without labels, captures coherent structures and realistic textures, demonstrating that prototypes learned online can effectively guide the denoiser. s-PDM further enhances sharpness and class alignment by leveraging label-informed prototypes. Together, these results illustrate the central contribution of our work: \textit{learning prototypes jointly with the diffusion model provides strong semantic conditioning that improves both visual fidelity and structural coherence}.

\section{Conclusion}
We presented the \textit{Prototype Diffusion Model} (PDM) and its supervised variant \textit{s-PDM}, which integrate prototype learning directly into the diffusion process. By jointly updating prototypes with the denoiser, PDM provides label-free semantic guidance without external memory or retrieval, while s-PDM leverages class labels for improved FID/KID performance through stable, class-specific prototypes.\newline
Experiments demonstrate that prototype-guided conditioning consistently enhances generative fidelity: s-PDM achieves the strongest quantitative results, and PDM offers a competitive, annotation-free alternative. Ablations confirm the importance of aligning the number of prototypes with dataset semantics.\newline
Future work includes extending prototype-guided diffusion to \textbf{text-to-image} tasks, exploring semi- and weakly supervised variants, and developing hierarchical or adaptive prototypes for complex and evolving data distributions.

\bibliographystyle{IEEEtran}
\bibliography{sample-base}
\end{document}